\pdfoutput=1

\documentclass[11pt]{article}

\usepackage[final]{acl}

\usepackage{times}
\usepackage{latexsym}

\usepackage[T1]{fontenc}

\usepackage[utf8]{inputenc}

\usepackage{microtype}

\usepackage{inconsolata}

\usepackage{graphicx}

\usepackage{pifont}
\usepackage{amsmath}
\usepackage{graphicx}
\usepackage{booktabs}
\usepackage{tikz}
\usepackage{pgfplots} 
\usepackage{scalefnt} 
\usepackage{caption}
\usepackage{subcaption}
\usepackage{enumitem}
\usepackage{todonotes}
\usepackage{colortbl}
\usepackage{makecell}
\usepackage{multirow}
\usepackage{xcolor}         
\usepackage{spverbatim}

\usepackage{algorithm}
\usepackage{algpseudocode}

\usepackage{stfloats}
\usepackage{setspace}

\title{UnitCoder: Scalable Iterative Code Synthesis with Unit Test Guidance}

\author{
 \textbf{Yichuan Ma\textsuperscript{1,2}},
 \textbf{Yunfan Shao\textsuperscript{1,2}},
 \textbf{Peiji Li \textsuperscript{1,2}},
 \textbf{Demin Song\textsuperscript{2}},
\\
 \textbf{Qipeng Guo\textsuperscript{2}},
 \textbf{Linyang Li\textsuperscript{2}$^\dag$},
 \textbf{Xipeng Qiu\textsuperscript{1}},
 \textbf{Kai Chen\textsuperscript{2}},
\\
 \textsuperscript{1}School of Computer Science, Fudan University, Shanghai,
\\
 \textsuperscript{2}Shanghai AI Laboratory, Shanghai,
\\
 \small{
  \href{yichuanma24@m.fudan.edu.cn}{yichuanma24@m.fudan.edu.cn},\href{lilinyang@pjlab.org.cn}{lilinyang@pjlab.org.cn}
 }
}

\begin{document}
\maketitle

\def\thefootnote{$\dag$}\footnotetext{Corresponding Author.}\def\thefootnote{\arabic{footnote}}

\begin{abstract}

Large Language Models (LLMs) have demonstrated remarkable capabilities in various tasks, yet code generation remains a major challenge.
Current approaches for obtaining high-quality code data primarily focus on (i) collecting large-scale pre-training data and (ii) synthesizing instruction data through prompt engineering with powerful models. While pre-training data faces quality consistency issues, instruction-based synthesis suffers from limited instruction diversity and inherent biases of LLMs.
To address this gap, we introduce UnitCoder, a systematic pipeline leveraging model-generated unit tests to both guide and validate the code generation process.
Combined with large-scale package-based retrieval from pre-training corpus, we generate a dataset of 500K+ verifiable programs containing diverse API calls.
Evaluations on multiple Python benchmarks (BigCodeBench, HumanEval, MBPP) demonstrate that models fine-tuned on our synthetic data exhibit consistent performance improvements.
Notably, Llama3.1-8B and InternLM2.5-7B improve from 31\% and 28\% to 40\% and 39\% success rates on BigCodeBench, respectively.
Our work presents a scalable approach that leverages model-generated unit tests to guide the synthesis of high-quality code data from pre-training corpora, demonstrating the potential for producing diverse and high-quality post-training data at scale.
All code and data will be released\footnote{\url{https://github.com/}}.

\end{abstract}

\section{Introduction}
Large language models (LLMs) have demonstrated remarkable capabilities in code-related tasks, as evidenced by both general-purpose foundation models and specialized solutions. Leading foundation models like OpenAI O1\footnote{\url{https://openai.com/o1/}}, GPT-4~\cite{Achiam2023GPT4TR}, Claude\footnote{\url{https://www.anthropic.com/claude}}, and DeepSeek R1~\cite{deepseekai-R1} excel at code understanding and generation. Meanwhile, specialized models such as CodeLlama~\cite{Rozire2023CodeLO}, Deepseek-Coder~\cite{guo2024deepseekcoder} and Qwen-Coder~\cite{Hui2024Qwen25CoderTR} have emerged as powerful assistants specifically designed for software development. 


Despite the success of LLMs in coding tasks, acquiring high-quality code data still remains a fundamental challenge. Current approaches to code data acquisition primarily follow two paths. The first focuses on large-scale data mining and cleaning, collecting comprehensive code repositories from various sources such as GitHub and Stack Overflow\footnote{\url{https://stackoverflow.com/}}. During the training phase, these large-scale pre-training code datasets are usually combined with high-quality instruction tuning data 
to comprehensively enhance the model's code understanding and generation capabilities~\cite{guo2024deepseekcoder, yang2024qwen2, Hui2024Qwen25CoderTR}. The second approach focuses on prompt engineering methods, utilizing powerful code models to synthesize data based on instructions~\cite{luo2023wizardcoder, Huang2023AgentCoderMC, Yu2023WaveCoderWA}.



However, both approaches face limitations. Large-scale pre-training approaches provide diversity but suffer from inconsistent quality and coding styles in their source data. Meanwhile, prompt engineering-based methods, exemplified by WizardCoder~\cite{luo2023wizardcoder} and AgentCoder~\cite{Huang2023AgentCoderMC}, concentrate on instruction augmentation and task-specific solutions using powerful models. Although effective in targeted scenarios, these methods are limited by LLM biases and the constraints of their original instruction sets, potentially lacking the diversity required for broader applications.

Therefore, it is a natural idea to design an LLM-based code synthesis method that directly leverages pre-training corpora, utilizing both the inherent diversity of large-scale data and LLMs' code capabilities. While prior works like OSS-Instruct~\cite{oss-instruct} have demonstrated the feasibility of corpus-based instruction synthesis, existing approaches often rely heavily on powerful models' capabilities.

To address this limitation, we introduce code executability as an objective quality metric to guide the synthesis process. Morevover, for code involving complex API interactions and exception handling, simple input-output validation may be insufficient. This observation leads us to leverage unit tests as a more comprehensive evaluation metric. Based on these insights, we introduce \textbf{UnitCoder}, a scalable code synthesis pipeline built upon pre-training code corpora, which employs model-generated unit tests for both guidance and validation of synthesis.

The UnitCoder framework consists of three key components: (i) Data Preparation, (ii) Fix and Refine Flow and (iii) Post-Train. In the first stage, we utilize the AST (Abstract Syntax Tree) parsing tool\footnote{\url{https://docs.python.org/3/library/ast.html}} to extract syntactically valid functions from the pre-training corpora. Additionally, we develop a unit test generator fine-tuned on human-written Python test cases, capable of validating complex API calls and boundary condition handling, enabling us to filter executable functions from the parsed code. In the second stage, for functions failing the unit tests, we employ LLMs to iteratively debug and modify code based on failure traces. Once the code passes the unit test, we introduce a refine agent to maintain consistent code style and improve readability without altering functionality. In the final stage, we conduct post-training of the base model using the generated code dataset. All agents in our experiment are implemented using open-source large language models, including Llama3-70B and Qwen2.5-72B. Utilizing this framework, we successfully synthesize over 500K executable code data along with the corresponding unit tests, covering over 370 diverse Python packages. 

We evaluate our approach through comprehensive experiments, using UnitCoder-generated data to fine-tune the Llama~\cite{dubey2024llama3} and InternLM~\cite{cai2024internlm2} series models. Fine-tuning with UnitCoder data improves model performance across all benchmarks, with the most noticeable improvement on BigCodeBench, where complex API interactions are required.

Our contributions can be summarized as follows:
\begin{itemize}
    \item We present UnitCoder, a scalable framework for synthesizing high-quality post-training code data from raw code corpora under unit test guidance. UnitCoder innovatively leverages code executability through unit tests as the primary guidance, ensuring the synthesis of high-quality data while preserving the original code functionality.
    \item We generate a dataset of 500K+ verifiable programs using UnitCoder. Extensive experiments demonstrate that our synthetic data consistently improves models' performance on code generation benchmarks, particularly in handling complex API interactions.
    \item We conduct comprehensive ablation studies to validate each component's necessity and analyze the relationships between data scale, diversity, and model performance, providing insights for scalable code synthesis.
\end{itemize}

\section{Related Work}

\begin{figure*}[h]
    \centering
    \includegraphics[width = 1\linewidth]{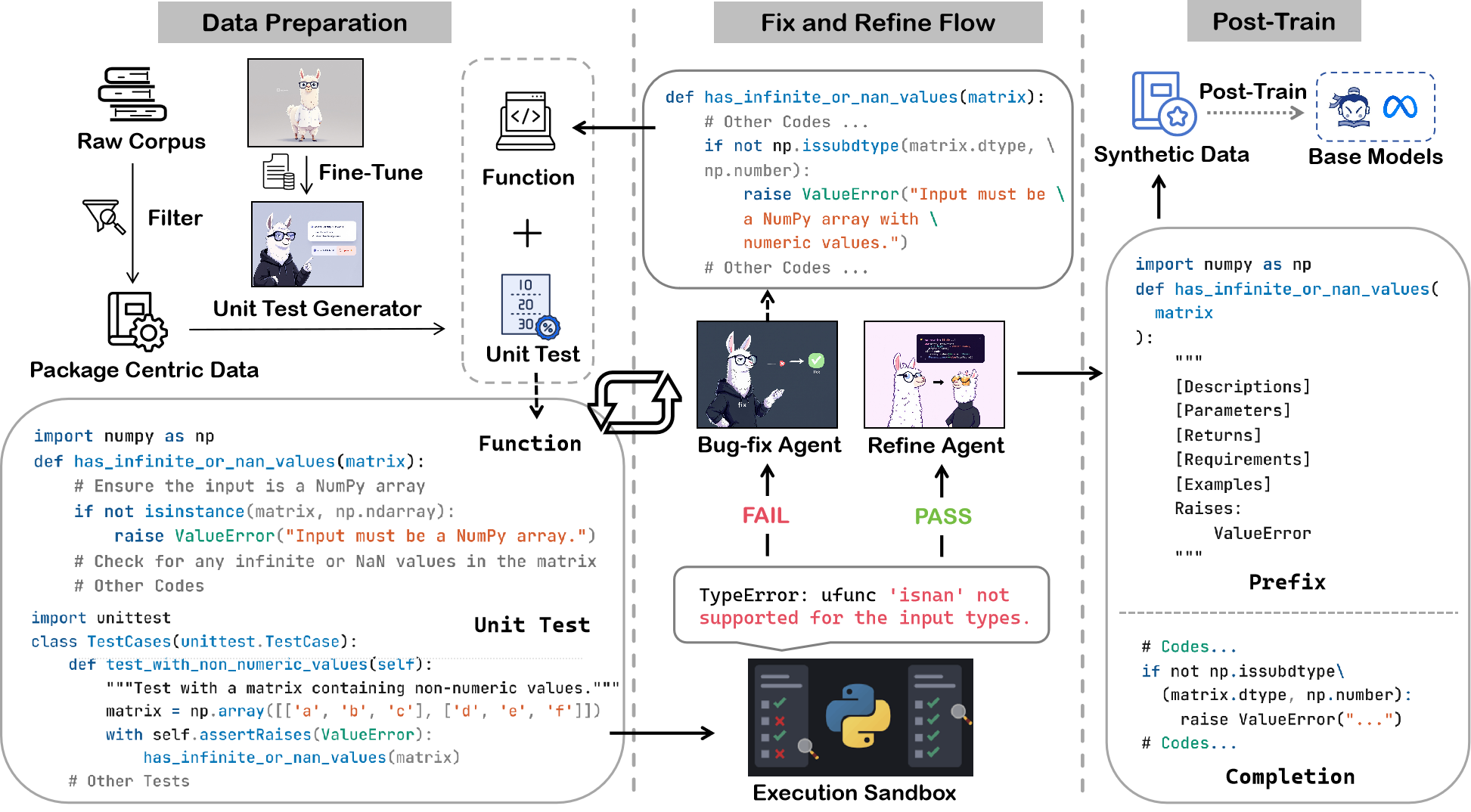}
    \caption{\textbf{The UnitCoder pipeline.} The pipeline consists of three main stages: (1) Data Preparation - filter package-centric data from raw code corpus and fine-tune a unit test generator to produce corresponding tests; (2) Fix and Refine Flow - execute function-test pairs in sandbox, iteratively fix failed cases via bug-fix agent, and refine successful code through refine agent; (3) Post-Train - construct prefix-completion pairs for post-training.}
    \label{fig:overall_framework}
\end{figure*}

\paragraph{Code LLMs}

Code LLM developments have progressed along two main directions: large-scale pre-training and specialized instruct-tuning.
Early works of code pre-training models include pioneering works like CodeX \cite{Chen2021EvaluatingLL}, CodeGen \cite{nijkamp2022codegen}, StarCoder~\cite{li2023starcoder} and CodeLlama \cite{Rozire2023CodeLO}.
Open-sourcing series such as Qwen~\cite{Bai2023QwenTR,Yang2024Qwen2TR} and Deepseek~\cite{deepseekai2024deepseekv2strongeconomicalefficient} build specialized code models as well, examplified by Qwen-Coder and Deepseek-Coder\cite{Hui2024Qwen25CoderTR, guo2024deepseekcoder}. 

Furthermore, leveraging language models for code data filtering provides a valuable quality supervision signal in data preparation. For example, WaveCoder~\cite{Yu2023WaveCoderWA} employs GPT-4 as a discriminator, while similar work such as Arctic-SnowCoder~\cite{Arctic-SnowCoder} explores the potential of BERT-based models for code data filtering.

In parallel, 
works represented by WizardCoder~\cite{luo2023wizardcoder} focus on enhancing instruction diversity through improved instruction engineering with powerful LLMs~\cite{Jiang2024ASO,Zan2023CanPL,Zhu2022ASO}.  
Additionally, research exemplified by AgentCoder~\cite{Huang2023AgentCoderMC} investigates prompt-based approaches that integrate test cases and multi-agent collaboration to improve coding performance~\cite{Huang2023AgentCoderMC, chen2022codet, islam2024mapcoder}.


\paragraph{LLM-Driven Unit Test Generation}

Software engineering is a major topic in coding and programming, with LLM-based unit test generation emerging as a promising direction. Large language models have demonstrated remarkable capabilities in this area, as exemplified by TestPilot~\cite{empirical-unittest}, which introduces a comprehensive framework for automated test generation using LLMs. 
Several other works further validate LLMs' effectiveness in generating high-quality test cases~\cite{chatunitest, chattester}.
Building upon these foundations, subsequent works continue to explore LLM-based unit test generation. Several works focus on improving metrics like coverage and accuracy~\cite{Achiam2023GPT4TR, codeaware_prompt, pizzorno2024coverup}. 
These works collectively demonstrate the potential of LLMs in advancing automated unit test generation practices.

\section{Method}

In this section, we present UnitCoder, a scalable code synthesis pipeline that leverages pre-training code corpora and employs model-generated unit tests for both synthesis guidance and quality validation. The complete framework is illustrated in Figure \ref{fig:overall_framework}.

The pipeline comprises three principal stages. In the first stage, we perform filtering of executable functions from a large-scale pre-trained code corpus. We then fine-tune a large language model to serve as our unit test generator, denoted as $\pi_{\theta_0}$.

In the second stage, we establish an iterative code improvement framework with two key components: (i) a debugging agent that identifies and fixes potential  defects in the original implementation through analysis of failed test cases and execution results, and (ii) a refinement agent that enhances code quality by adding docstrings and standardizing coding conventions once the code successfully passes the unit test.

In the post-training stage, we leverage the synthesized data to conduct post-training on open-source foundation models to validate the effectiveness of our approach.


\subsection{Data Preparation}

In the first stage of the UnitCoder pipeline, we filter executable function snippets from pre-training code corpus, and fine-tune a unit test generator to generate corresponding unit tests for the filtered functions.

\subsubsection{Package-based Function Extraction}
\label{sec:package retrieve}
We extract executable code snippets from pre-training corpus through a two-step process: First, we perform AST-based semantic analysis to identify syntactically valid function units. Then, we filter these functions based on a predefined list of common APIs to retain those with meaningful package imports. This process yields a subset $\mathcal{D}_{pkg}$ from the original dataset $\mathcal{D}$.


\subsubsection{Unit Test Generation}

To generate corresponding unit tests for the extracted code snippets, we fine-tune a unit test generator that thoroughly evaluates complex function implementations. The generator creates comprehensive test cases to verify function behavior across edge cases, error conditions, and intricate API interactions. This generator, denoted as $\pi_{\theta}$, is built upon Llama3-70B-Instruct and fine-tuned using high-quality function-test pairs. For each executable function $f_i \in \mathcal{D}_{pkg}$, $\pi_{\theta_0}$ generates a corresponding unit test $u_i$.

\subsection{Fix and Refine Flow}

In the second stage, we design an iterative code improvement framework based on unit test execution results. We utilize an open-source LLM to debug and fix code according to error traces, followed by a refinement step to ensure consistency in the quality of synthesized data. To ensure safe operation while processing code from unknown sources, we implement a security sandbox for code execution.

\subsubsection{Safety preparation}
For safe execution, we build a secure sandbox, which redirects potentially risky operations, including file system operations like directory creation and deletion.


\subsubsection{Iterative Code Improvement}
After implementing security measures, we pair and execute functions along with their corresponding unit tests to obtain initial execution results. Specifically, for each function $f_{i}^{0} \in \mathcal{D}_{pkg}$, its corresponding unit test is denoted as $u_i$. Functions that pass their unit tests are collected into a set $\mathcal{D}_{pass}$, while those that fail are placed in another set $\mathcal{D}_{curr}^{0}$ for subsequent iterative debugging. 


\begin{algorithm}[t]
\setstretch{0.9}
\caption{Code Improvement Pipeline}\label{alg:codeimprovement}
\begin{algorithmic}
\State $\mathcal{D}_{pass} \gets \emptyset$ \Comment{Repository of validated code}
\State $\mathcal{D}_{curr}^{0} \gets \emptyset$ \Comment{Queue of pending code}
\State $r=0$ \Comment{Current iteration counter}
\State $max\_round \in N$ \Comment{Maximum iteration limit}

\State \textbf{\textit{Phase 1: Unit Test Initialization}}
\For{each function $f_{i}^{r}$ in $\mathcal{D}_{p\_safe}$}
    \State Generate comprehensive test suite for $f_{i}^{r}$
    \State $u_{i} \gets \pi_{\theta_0}(f_{i}^{r})$
    \If{$f_{i}^{r}$ passes unit test $u_{i}$}
        \State Archive successfully validated code
        \State $\mathcal{D}_{pass} \gets \mathcal{D}_{pass} \bigcup \{(f_{i}^{r}, u_{i}, r_{i}^{r})\}$ 
    \Else
        \State Record execution diagnostics
        \State $\mathcal{D}_{curr}^{r} \gets  \mathcal{D}_{curr}^{r} \bigcup \{(f_{i}^{r}, u_{i}, r_{i}^{r})\}$
    \EndIf
\EndFor

\State \textbf{\textit{Phase 2: Iterative Code Improvement}}
\State $r=1$
\While{$r \le max\_round$}
    \State $\mathcal{D}_{curr}^{r} \gets \emptyset$
    \For{$(f_{i}^{r-1}, u_{i}, r_{i}^{r-1} ) \in \mathcal{D}_{curr}^{r-1}$}
        \State Apply improvement to $f_{i}^{r-1}$
        \State $f_{i}^{r} \gets \pi_{\theta_1}(f_{i}^{r-1}, u_{i}, r_{i}^{r-1})$
        \If{$f_{i}^{r}$ passes unit test $u_{i}$}
            \State $\mathcal{D}_{pass} \gets D_{pass} \bigcup \{(f_{i}^{r}, u_{i}, r_{i}^{r})\}$ 
        \Else
            \State $\mathcal{D}_{curr}^{r} \gets D_{curr}^{r} \bigcup \{(f_{i}^{r}, u_{i}, r_{i}^{r})\}$ 
        \EndIf
        \State Collected functions that pass the unit tests.
    \EndFor
    \State $r$ ++
\EndWhile
\end{algorithmic}
\end{algorithm}

Now that we collect the failed code snippets, the related unit tests, and the execution results, the bug-fix agent $\pi_{\theta_1}$ is employed to iteratively revise the failed codes. The complete process of code improvement is detailed in Algorithm \ref{alg:codeimprovement}.

For the $r$-th iteration, let $\mathcal{D}_{curr}^{r-1}$ denote the collection of failed functions from round $r-1$. For each function $f_{i}^{r-1} \in \mathcal{D}_{curr}^{r-1}$ with its corresponding unit test $u_i$ and execution result $r_i^{r-1}$, the revision step can be formulated as:
\begin{equation*}
    f_{i}^{r} = \pi_{\theta_1}(f_{i}^{r-1}, u_{i}, r_{i}^{r-1})
\end{equation*}

The revised function $f_{i}^{r}$ is then evaluated using its corresponding unit test $u_{i}$. Functions that pass the unit tests are collected into $\mathcal{D}_{pass}$, while the failed ones are collected into $\mathcal{D}_{curr}^{r}$ for the next iteration. This iterative revision process continues until reaching the maximum iteration bound, accumulating all successfully fixed functions throughout the iterations.





\subsubsection{Code Refinement}


Through the iterative improvement process, we have constructed $\mathcal{D}_{pass}$, a collection of validated functions that pass their corresponding unit tests. Given that the original code corpus $\mathcal{D}$ is sourced from diverse repositories, it is necessary to normalize the coding style to ensure consistency in the synthetic data quality.

To address this requirement, we introduce a refine agent $\pi_{\theta_2}$ that enhances code readability in three aspects: (i) generating informative docstrings in natural language, (ii) adding explanatory inline comments at key code sections, and (iii) maintaining consistent coding style conventions.


\subsection{Post-Train}

In the third stage, we construct the post-training dataset $\mathcal{D}_{Unit}$ by reformulating the validated functions into supervised learning samples. Each training sample is structured as a pair, where the input consists of the import statements, function signature, and descriptive docstring, and the output contains the complete function implementation. 

Subsequently, we conduct post-training on open-source foundation models using our synthetic dataset. Experimental results demonstrate both the diversity and high quality of our synthetic data, validating the effectiveness of the UnitCoder pipeline.


\section{Experiment}
In this section, we first briefly introduce the experimental setups, then discuss the experimental results, thoroughly demonstrating and validating the effectiveness of the UnitCoder pipeline.

\subsection{Experimental Setups}

\paragraph{Training Setup}
For the unit test generator $\pi_{\theta_0}$, we employ Llama3-70B-Instruct as the foundation model. The post-training experiments are conducted on InternLM-2.5-7B and Llama-3.1-8B. We also perform ablation studies on the InternLM series to examine the impact of model scale. Both fine-tuning and post-training processes run for 1 epoch, with learning rates following a linear warmup and cosine decay schedule (1e-5 to 3e-6) and a maximum context window of 4096 tokens. The training utilizes A800 GPUs, with 64 GPUs for Llama3-70B-Instruct fine-tuning and 16 GPUs for smaller models.

\paragraph{Evaluation Setup}


We evaluate our post-trained models on three standard code benchmarks: BigCodeBench, HumanEval, and MBPP~\cite{zhuo2024bigcodebench, chen2021evaluating-humaneval, austin2021program-mbpp}, based on the OpenCompass framework~\cite{2023opencompass}. The evaluation employs a 3-shot strategy for HumanEval and MBPP, while using complete mode for BigCodeBench. For unit test generator evaluation, we use solutions from HumanEval and MBPP as inputs to assess the accuracy of generated unit tests.

\paragraph{Unit Test Generator Setup}

We fine-tune the unit test generator $\pi_{\theta_0}$ based on Llama3-70B-Instruct. The fine-tuning data consists of unit test-function pairs from BigCodeBench, comprising 1140 functions with rich API calls and their corresponding human-written unit tests. To prevent evaluation set leakage, during supervised fine-tuning (SFT), we mask the original function when computing the loss.


\begin{table*}[t]
  \centering
  \resizebox{0.9\textwidth}{!}{
    \begin{tabular}{lcccc}
    \toprule
      & \textbf{HumanEval} & \textbf{MBPP} & \textbf{BigCodeBench} & \textbf{BigCodeBench-Hard} \\
    \midrule
    Llama3.1-8B & 36.6 & 58.8 & 31.0 & 5.4 \\
    \multicolumn{1}{c}{\textbf{+UnitCoder}}  & \textbf{61.0} & \textbf{63.4} & \textbf{40.4} & \textbf{14.2} \\
    \midrule
    InternLM2.5-7B &  65.2 & 60.3 & 27.9 & 10.1  \\
    \multicolumn{1}{c}{\textbf{+UnitCoder}} & \textbf{67.1} & \textbf{66.2} & \textbf{39.3} & \textbf{17.6} \\
    \midrule
    InternLM2.5-7B-Base & 41.5 & 57.6 & 28.3 & 7.4 \\
   \multicolumn{1}{c}{\textbf{+UnitCoder}}  & \textbf{62.2} & \textbf{65.8} & \textbf{41.6} & \textbf{14.9} \\
    \bottomrule
    \end{tabular}%
    }
  \caption{Performance of base models post-trained with UnitCoder synthetic data. Results of BigcodeBench are tested under "complete" mode.}
  \label{tab:FULL results}%
  \vspace{-10pt}
\end{table*}%

\begin{table}[t]
  \centering
  \resizebox{0.9\linewidth}{!}{
    \begin{tabular}{l|cc}
    \toprule
     Models & \textbf{BCB} & \textbf{BCB-Hard} \\
    \midrule
    \multicolumn{3}{c}{\cellcolor{gray!20} Base Models (7B size)} \\ 
    \midrule
    Llama3.1-8B & \underline{31.0} & 5.4  \\
    InternLM2.5-7B & 27.9 & \underline{10.1} \\ 
    InternLM2.5-7B-Base & 28.3 & 7.4 \\ 
    \midrule
    \multicolumn{3}{c}{\cellcolor{gray!20} Chat Models (7B size)} \\ 
    \midrule
    Mistral-7B-Instruct-v0.3 & 25.7 & 6.8 \\ 
    Qwen2.5-7B-Instruct & \underline{42.4} & \underline{14.2} \\ 
    Llam3.1-8B-Instruct & 39.6 & 10.8 \\ 
    InternLM2.5-7B-Chat & 32.9 & 5.4 \\ 
    \midrule
    \multicolumn{3}{c}{\cellcolor{gray!20} Code LLMs (7B size)} \\ 
    \midrule
    CodeLlama-7B-Instruct & 27.3 & 4.1 \\ 
    Deepseek-Coder-6.7B & 40.4 & 11.5 \\ 
    CodeQwen1.5-7B & 43.4 & 14.8 \\ 
    Qwen2.5-Coder-7B & \underline{\textbf{45.3}} & \underline{15.9} \\ 
    \midrule
    \multicolumn{3}{c}{\cellcolor{gray!20} Ours (7B size)} \\ 
    \midrule
    Llam3.1-8B+$\mathcal{D}_{Unit}$ & 40.4 & 14.2 \\
    InternLM2.5-7B-Base+$\mathcal{D}_{Unit}$ & \underline{41.6} & 14.8 \\ 
    InternLM2.5-7B+$\mathcal{D}_{Unit}$ & 39.3 & \underline{\textbf{17.6}} \\
    \bottomrule
    
    \end{tabular}%
    }
  \caption{Performance comparison between our proposed method and existing models on BigCodeBench (BCB) and BigCodeBench-Hard(BCB-Hard).}
  \label{tab:in-domain-models}%
  \vspace{-10pt}
\end{table}%

\begin{table}[t]
  \centering
  \resizebox{0.8\linewidth}{!}{
    \begin{tabular}{lcc}
    \toprule
     Method & \textbf{BCB} & \textbf{BCB-Hard} \\
    \midrule
    \textbf{Llama3.1-8B} & 31.0 & 5.4  \\
    \quad +Evol & 26.2 & 6.1 \\
    \quad +OSS & 27.5 & 6.8 \\
    \quad +OpenCoder & 32.3 & 10.1 \\
    
    \quad \textbf{+UnitCoder} & \textbf{40.4} & \textbf{14.2} \\
    
    \midrule
    \textbf{InternLM2.5-7B} & 27.9 & 10.1 \\ 
    \quad +Evol & 21.1 & 4.1 \\
    \quad +OSS & 22.8 & 5.4 \\
    \quad +OpenCoder & 28.2 & 8.8 \\
    \quad \textbf{+UnitCoder} & \textbf{39.3} & \textbf{17.6} \\
    
    \bottomrule
    
    \end{tabular}%
    }
  \caption{Performance comparison between UnitCoder dataset and other synthetic datasets on BigCodeBench (BCB) and BigCodeBench-Hard(BCB-Hard). Evol, OpenCoder and OSS-Instruct refer to Evol-codealpaca-v1, OpenCoder-SFT-Stage-1 and OSS-Instruct-75K datasets, respectively.}
  \label{tab:in-domain-datasets}%
  \vspace{-10pt}
\end{table}%

\paragraph{Data Preparation}


In the UnitCoder pipeline, our pre-training code corpus primarily comes from The Stack pre-training dataset, where we have already performed data deduplication with evaluation benchmarks (e.g., HumanEval, MBPP, BigCodeBench, etc.). Additionally, we utilize an SFT dataset from WizardCoder~\cite{luo2023wizardcoder}, which serves as complementary data mixed with our synthetic data during the post-training stage, in order to maintains instruction-following capabilities.

To validate UnitCoder's effectiveness in complex API interactions, we compare against several synthetic datasets: \textbf{OSS-Instruct}~\cite{oss-instruct}: A dataset of 75,000 instruction-code pairs synthesized from raw code.
\textbf{OpenCoder-SFT-Stage-1}~\cite{Huang2024OpenCoderTO}: A collection of 4.2M question-answer pairs spanning diverse computer science domains, generated from general code corpora.
\textbf{Evol-codealpaca-v1}~\cite{luo2023wizardcoder}: A dataset of 110K instruction pairs created by augmenting instructions using GPT-4.

\subsection{Post-Training Performance Analysis}



Table \ref{tab:FULL results} demonstrates the effectiveness of UnitCoder in enhancing LLMs' code capabilities. Our post-training approach combines synthetic data and SFT data, and achieves significant improvements across all base models.

On the BigCodeBench benchmark, which evaluates package calling capabilities in complex scenarios, UnitCoder significantly improves the performance of multiple base models: InternLM 2.5-7B's accuracy increases from 27.9\% to 39.3\%, InternLM 2.5-7B-base from 28.3\% to 41.6\%, and Llama3.1-8B from 31.0\% to 40.4\%. Furthermore, post-trained base models demonstrate consistent improvements across other code benchmarks, including HumanEval and MBPP. These comprehensive performance gains across multiple benchmarks validate the effectiveness of the UnitCoder approach.

\subsection{Analysis of Comparative Experiments}
We conduct extensive comparative experiments on BigCodeBench to comprehensively evaluate the effectiveness of our approach on complex API invocation tasks. Table \ref{tab:in-domain-models} presents comparisons among 7B-scale models, including base models, instruction-tuned models, and code-specialized models. Our method achieves comparable performance to leading instruction-tuned models, and significantly outperforms mainstream pre-trained models. Notably, on BigCodeBench-Hard, which evaluates complex API composition capabilities, our approach matches or even exceeds the performance of code-specialized models of similar size.

To further evaluate the effectiveness of our approach, we conducted controlled experiments by fine-tuning the same base model with different training datasets, as shown in Table \ref{tab:in-domain-datasets}. The results demonstrate that our method achieves the most significant performance improvements on BigCodeBench among all compared approaches. This superior performance on API-related tasks clearly validates the quality of our synthetic dataset and the effectiveness of the UnitCoder framework, especially considering these improvements were achieved with a relatively compact dataset.



\begin{figure*}[t]
    \centering
    \includegraphics[width=0.9\linewidth]{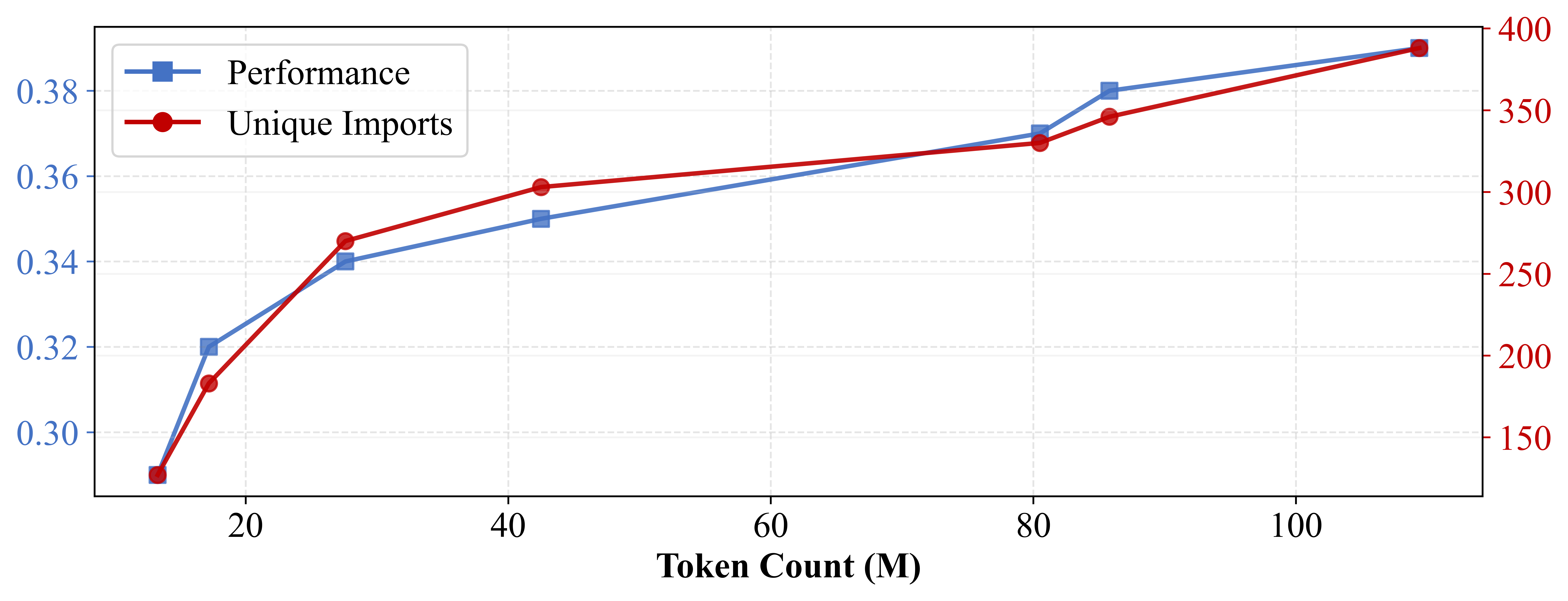}
    \caption{Scaling Effects of Synthetic Data: As the scale of synthetic data (measured in tokens) increases, we observe a corresponding growth in both the diversity of unique packages in synthetic data and InternLM2.5-7B's performance on BigCodeBench after post-training.}
    \label{fig:token-pkg-performance}
\end{figure*}

\begin{table*}[t]
  \centering
  \resizebox{0.9\textwidth}{!}{
    \begin{tabular}{l|cccc}
    \toprule
     \textbf{Method} & \textbf{HumanEval} & \textbf{MBPP} & \textbf{BigCodeBench} & \textbf{BigCodeBench-Hard} \\
    \midrule
    Base Model & \underline{65.2} & 60.3 & 27.9 & 10.1 \\
    + General Code & 58.5 & 61.1 & 29.4 & 7.4 \\
    + $\mathcal{D}_{pkg}$ & 50.6 & 54.5 & 29.7 & 4.1 \\
    + General Code + $\mathcal{D}_{pkg}$ & 61.0 & \underline{62.3} & 31.1 & 6.1 \\
    + General Code + $\mathcal{D}_{pass}$ & 61.6 & 61.1 & \underline{35.2} & \underline{13.5} \\
    \midrule
    + General Code + $\mathcal{D}_{Unit}$\textbf{(Ours)} & \textbf{67.1} & \textbf{66.2} & \textbf{39.3} & \textbf{17.6} \\
    \bottomrule
    \end{tabular}%
    }
    \caption{Ablation study of the UnitCoder pipeline, showing performance comparison of InternLM-2.5-7B under different training configurations. The evaluation demonstrates the impact of various training data combinations: general code data (General SFT dataset), $\mathcal{D}_{pkg}$(package-centric subset without verification), $\mathcal{D}_{pass}$(Verified dataset without refine), and $\mathcal{D}_{Unit}$(verified and refined data generated through the UnitCoder pipeline).}
  \label{tab:Ablation}%
  \vspace{-10pt}
\end{table*}

\begin{figure}[!t]
    \centering
    \includegraphics[width=1.0\linewidth]{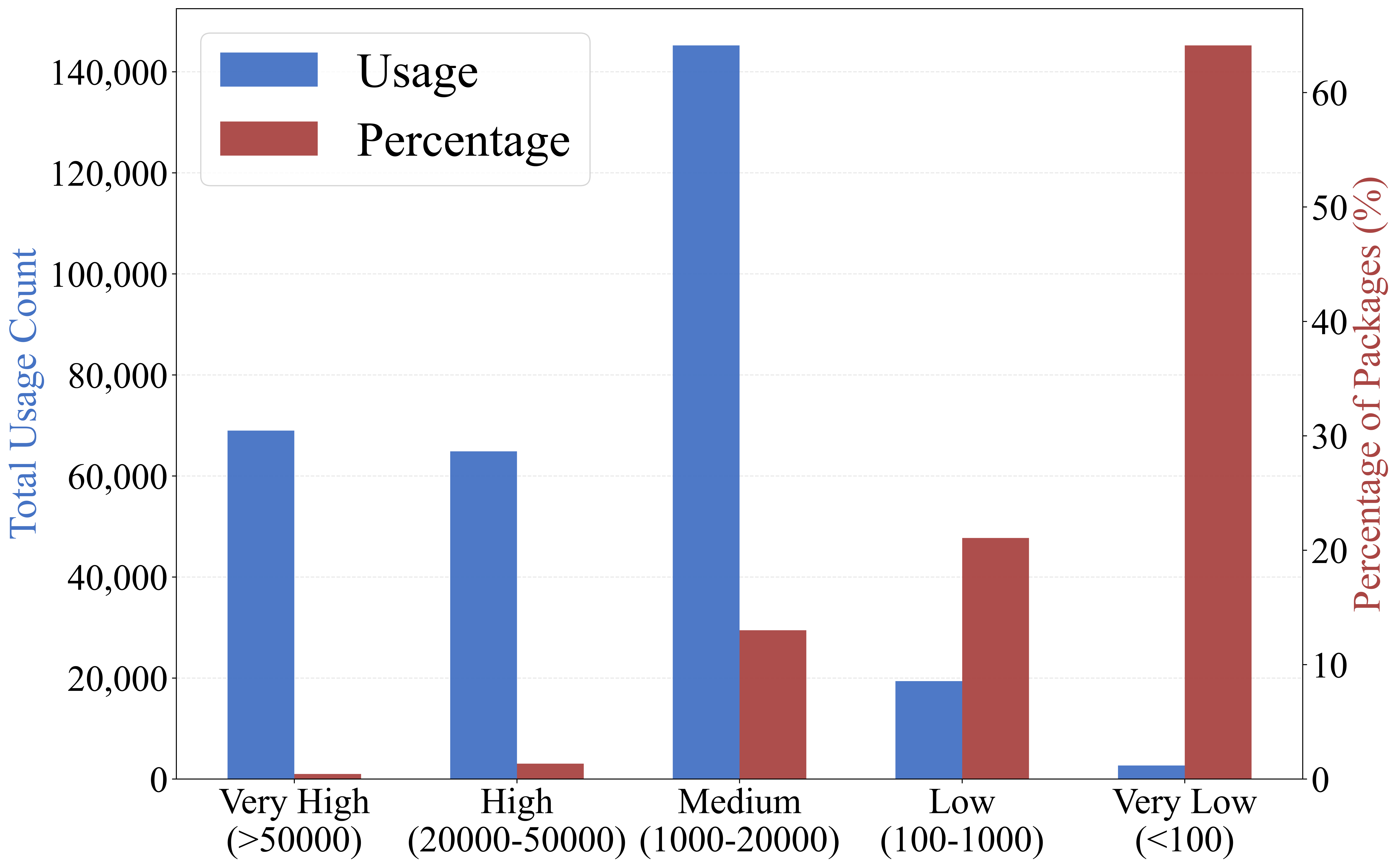}
    \caption{The distribution of packages in filtered code data, grouped by usage frequency. Usage represents the frequency of package imports, and Percentage shows the percentage of package types within each frequency group relative to the total number. }
    \label{fig:pkg_distribution}
\end{figure}





\subsection{Scaling Effects of Synthetic Data}

For investigating the impact of synthetic data scale on model performance, we conduct a series of controlled experiments with increasing data size. We first analyze the occurrence distribution of different APIs in the original code corpus, as shown in Figure \ref{fig:pkg_distribution}. The results demonstrate a distinct long-tail distribution, where a small number of frequently used packages account for the majority of invocations, while most packages exhibit relatively low occurrence frequencies.

For packages with lower occurrence frequencies, UnitCoder demonstrates effective verification and synthesis of high-quality data. As shown in Figure \ref{fig:token-pkg-performance}, the expansion of synthetic data scale leads to two significant improvements: First, it enables the capture of a broader spectrum of API call patterns, particularly those that appear infrequently in the original corpus. Second, it contributes to consistent performance improvements on BigCodebench, where complex package usage is needed.


\subsection{Ablation Studies}

In this section, we present ablation studies to comprehensively evaluate the key components of our UnitCoder pipeline. Our experiments address three critical research questions: (i) the necessity of the iterative verification process, (ii)  the impact of the refine agent, and (iii) the consistency of synthetic data's effectiveness across different model scales. 
Our experimental results are presented in Table~\ref{tab:Ablation} and Table~\ref{tab:Ablation scale}.

\begin{table}[t]
  \centering
  \resizebox{1\linewidth}{!}{
    \begin{tabular}{lcc}
    \toprule
     \textbf{} & \textbf{BigCodeBench} & \textbf{BigCodeBench-Hard} \\
    \midrule
    InternLM2.5-1.8B & 14.7 & 2.0 \\
    \multicolumn{1}{c}{\textbf{+UnitCoder}} & \textbf{19.6}  & \textbf{4.1} \\
    \midrule
    InternLM2.5-7B & 27.9 & 10.1 \\
    \multicolumn{1}{c}{\textbf{+UnitCoder}} & \textbf{39.3} & \textbf{17.6} \\
    \midrule
    InternLM2.5-20B & 41.1 & 14.2 \\
    \multicolumn{1}{c}{\textbf{+UnitCoder}} & \textbf{44.6} & \textbf{22.3} \\
    \bottomrule
    \end{tabular}%
    }
  \caption{Abaltion study on model scale.}
  \label{tab:Ablation scale}%
  \vspace{-10pt}
\end{table}

\label{sec:ablation study}


\paragraph{RQ1: How essential is the verification pipeline?} We first evaluate the effectiveness of the iterative code improvement module. As reported in Table~\ref{tab:Ablation}, we first compare two experimental settings: (i) fine-tuning with general SFT data alone, and (ii) fine-tuning with a combination of SFT data and unit-test verified data ($\mathcal{D}_{pass}$). 

Results show that the verification process brings substantial performance gains across all benchmarks, with the most notable improvement observed on BigCodeBench where the pass rate increases from 29.4\% to 35.2\%. This demonstrates the critical role of verification in enhancing model performance, even before applying subsequent refinement steps.

To further examine whether similar performance gains could be achieved without verification, we conduct a comparative experiment using unverified package-centric data ($\mathcal{D}_{pkg}$) combined with SFT data. The results show that our verification process yields a 4\% performance improvement on BigCodeBench and a substantial 7\% gain on BigCodeBench-Hard. These performance gains clearly demonstrate that the verification process is an irreplacable component of our pipeline. 

\paragraph{RQ2: Is code refinement necessary?} Following our verification pipeline analysis, we investigate the effectiveness of the refinement process through ablation studies in Table~\ref{tab:Ablation}. We compare two experimental settings: (i) fine-tuning with a combination of general SFT data and UnitCoder-synthesized data, and (ii) fine-tuning with SFT data combined with verified-only data ($\mathcal{D}_{pass}$). Our results show consistent performance improvements across all benchmarks after applying the refinement process.

Notably, compared to the $\mathcal{D}_{pass}$ mixture method, we observe particularly significant improvements on HumanEval and MBPP benchmarks, which focus less on complex package interactions. These results indicate that the refinement step comprehensively enhances the quality of synthetic data, enabling the model to better learn fundamental coding capabilities from the original codebase.

\paragraph{RQ3: Does the method work on different model scales?}
To further investigate the impact of our synthetic data, we trained models from the InternLM2.5 series across different scales. Table \ref{tab:Ablation scale} reports our results, demonstrating consistent performance improvements across all model sizes. 
Specifically, on BigCodeBench, the 1.8B variant improves from 14.6\% to 19.6\%, the 7B variant from 27.9\% to 39.3\%, and the 20B variant from 41.1\% to 44.6\%. Furthermore, on BigCodeBench-Hard, our synthetic data brings an average improvement of approximately 6\% across all model scales. 
These results highlight how effectively the synthetic data enhances performance on package-related coding tasks across various model sizes.

\subsection{Unit Test Generator Evaluation}
To assess our unit test generator, we conduct experiments using canonical solutions from HumanEval and MBPP as input functions, and evaluate whether the generated unit tests can effectively validate these functions. Results in Table~\ref{tab:unit test generator result} show that our generator produces tests with both high accuracy and comprehensive code coverage, demonstrating its reliability in providing meaningful verification for most code samples.

\begin{table}[t]
  \centering
  \resizebox{0.35\textwidth}{!}{
    \begin{tabular}{l|cc}
    \toprule
     Benchmark & \textbf{Accuracy} & \textbf{Coverage} \\
    \midrule
    HumanEval & 80.4 & 96.9  \\
    MBPP & 84.2 & 92.5 \\
    \bottomrule
    \end{tabular}%
    }
  \caption{Unit test generator evaluation}
  \label{tab:unit test generator result}%
  \vspace{-10pt}
\end{table}%

\section{Conclusion}
We present UnitCoder, a scalable framework for synthesizing high-quality post-training code data from raw code corpora under unit test guidance. The framework innovatively leverages code executability through unit tests as the primary guidance, ensuring the synthesis of high-quality data while preserving the original code functionality. By synthesizing a dataset of over 500K verifiable programs, we demonstrate through extensive experiments that our synthetic data consistently improves models' performance on code generation benchmarks, particularly in handling complex API interactions. Through comprehensive ablation studies, we validate each component's necessity and analyze the relationships between data scale, diversity, and model performance, providing valuable insights for scalable code synthesis. We believe UnitCoder demonstrates an effective approach for scalable, high-quality code data synthesis, providing valuable insights for future research in LLM-based code generation.

\section*{Limitations}
Despite the demonstrated effectiveness of UnitCoder, our approach has several limitations that warrant discussion.
First, while UnitCoder shows promising results with our current unit test generator, utilizing more advanced models could potentially improve synthesis quality. The trade-off between model capabilities and computational efficiency requires further investigation.
Second, our framework is currently limited to Python code synthesis. Extending UnitCoder to multiple programming languages would help validate its generalizability across different development contexts.
Third, although our method handles function-level synthesis effectively, scaling to project-level code generation remains challenging and requires further research.

\bibliography{custom}

\appendix

\section{Demonstration of the Pipeline}
\label{sec:appendix}

In this section, we demonstrate our synthetic pipeline through several examples. The demonstrations include:
\begin{itemize}
    \item A demonstration of unit test generation, showing the original function, its generated unit tests, and the test execution output format.
    \item The bug-fixing process, demonstrating how the system handles failed test cases.
    \item The complete refined function after all iterations and improvements.
    \item Prompts for the unit test generator and the agents.
\end{itemize}

\begin{table*}[htbp]
    \centering
    \small
    \begin{tabular}{l}
    \toprule
    \textbf{Unit Test Demo} \\
    \midrule
    \begin{minipage}{0.8\textwidth}
\begin{spverbatim}
Function:
import numpy as np

def drawWeights(size, distribution):
    # Validate the size parameter
    if not isinstance(size, (int, tuple)) or (isinstance(size, int) and size <= 0):
        raise ValueError("Size must be a positive integer or a tuple of positive integers.")
    
    # Validate the distribution parameter
    if distribution not in ['lognormal', 'normal', 'uniform']:
        raise ValueError("Distribution must be one of 'lognormal', 'normal', or 'uniform'.")
    
    weights = None
    if distribution == 'lognormal':
        hyp = 1.0
        m = np.log(0.2) + hyp
        s = hyp
        weights = (np.random.lognormal(m, s, size) * (255 / 20.0)).astype(int)
    elif distribution == 'normal':
        m = 10
        s = 5
        weights = np.random.normal(m, s, size).astype(int)
    elif distribution == 'uniform':
        weights = np.random.uniform(0, 255, size).astype(int)
    
    return weights
\end{spverbatim}\end{minipage}\\
\midrule
\begin{minipage}{0.8\textwidth}\begin{spverbatim}
Unit Test:
import unittest
import numpy as np
class TestCases(unittest.TestCase):
    def test_lognormal_weights(self):
        np.random.seed(42)
        weights = drawWeights(10, 'lognormal')
        self.assertIsInstance(weights, np.ndarray)
        self.assertEqual(len(weights), 10)
        self.assertTrue(all(0 <= w <= 255 for w in weights))
    def test_normal_weights(self):
        np.random.seed(42)
        weights = drawWeights(10, 'normal')
        self.assertIsInstance(weights, np.ndarray)
        self.assertEqual(len(weights), 10)
        self.assertTrue(all(-5 <= w <= 20 for w in weights))
    def test_uniform_weights(self):
        np.random.seed(42)
        weights = drawWeights(10, 'uniform')
        self.assertIsInstance(weights, np.ndarray)
        self.assertEqual(len(weights), 10)
        self.assertTrue(all(0 <= w <= 255 for w in weights))
    def test_invalid_size(self):
        with self.assertRaises(ValueError):
            drawWeights(-1, 'lognormal')
    def test_invalid_distribution(self):
        with self.assertRaises(ValueError):
            drawWeights(10, 'invalid_distribution')
\end{spverbatim}\end{minipage}\\
\midrule
\begin{minipage}{0.8\textwidth}\begin{spverbatim}
Execution Result: 
["pass",{}]
\end{spverbatim}
        \end{minipage}\\
    \bottomrule
    \end{tabular}
    \caption{Demo of a function, its corresponding unit test, and the execution result.}
    \label{tab:appendix-unittest-demo}
\end{table*}

\begin{table*}[htbp]
    \centering
    \small
    \begin{tabular}{l}
    \toprule
    \textbf{Debug Demo} \\
    \midrule
    \begin{minipage}{0.8\textwidth}
\begin{spverbatim}
Failed Test Case:
def test_data_file_with_non_image_entries(self):
    # Create a data file with non-image entries
    mixed_data_file_path = os.path.join(self.test_dir, "mixed_data.txt")
    with open(mixed_data_file_path, 'wt') as f:
        f.write("input_image3.png annotation_image3.png\n")
        f.write("non_image_data.txt\n")
    # Try to load the mixed data file
    with self.assertRaises(ValueError):
        _load_data_files(mixed_data_file_path)
\end{spverbatim}\end{minipage}\\
\midrule
\begin{minipage}{0.8\textwidth}\begin{spverbatim}
Original Execution Result:
[
    "fail",
    {
      "test_data_file_with_non_image_entries": "Traceback (most recent call last):\n  File \"__test__.py\", line 140, in test_data_file_with_non_image_entries\nAssertionError: ValueError not raised\n",
      "test_invalid_data_file_format": "Traceback (most recent call last):\n  File \"__test__.py\", line 117, in test_invalid_data_file_format\nAssertionError: ValueError not raised\n"
    }
]
\end{spverbatim}\end{minipage}\\
\midrule
\begin{minipage}{0.8\textwidth}\begin{spverbatim}
Fix Bug Accorrdingly:
import numpy as np
import os
import logging
# Configure logging
logging.basicConfig(level=logging.INFO, format='%(asctime)s - %(levelname)s - %(message)s')
def _load_data_files(data_file_path):
    input_path_list = []
    annotation_path_list = []
    # Check if the data file exists
    if not os.path.exists(data_file_path):
        logging.error(f"Data file not found: {data_file_path}")
        raise FileNotFoundError(f"Data file not found: {data_file_path}")
    data_folder_path = os.path.dirname(os.path.abspath(data_file_path))
    try:
        with open(data_file_path, 'rt') as f:
            for line in f:
                # Split the line into input and annotation paths
                parts = line.strip().split(' ')
                if len(parts) != 2:
                    logging.warning(f"Skipping invalid line: {line.strip()}")
                    raise ValueError(f"Invalid line format: {line.strip()}")
                input_path, annotation_path = parts
                # Construct full paths
                full_input_path = os.path.join(data_folder_path, input_path)
                full_annotation_path = os.path.join(data_folder_path, annotation_path)
                # Validate paths
                if not os.path.exists(full_input_path) or not os.path.exists(full_annotation_path):
                    logging.warning(f"Skipping non-existent paths: {full_input_path} or {full_annotation_path}")
                    raise ValueError(f"Non-existent paths: {full_input_path} or {full_annotation_path}")  ##### ValueError raise added here ##### 
                input_path_list.append(full_input_path)
                annotation_path_list.append(full_annotation_path)
    except Exception as e:
        logging.error(f"Error reading data file: {e}")
        raise
    return np.array(input_path_list), np.array(annotation_path_list)

\end{spverbatim}
        \end{minipage}\\
    \bottomrule
    \end{tabular}
    \caption{Demo of the code enhancement pipeline.}
    \label{tab:appendix-generator-prompt}
\end{table*}

\begin{table*}[htbp]
    \centering
    \small
    \begin{tabular}{l}
    \toprule
    \textbf{Refine Demo} \\
    \midrule
    \begin{minipage}{0.8\textwidth}
\begin{spverbatim}
import numpy as np

def get_var(data):
    """
    Calculates the variance of a given list of numbers.
    
    Parameters:
    - data (list of float or int): A list of numerical values for which to calculate the variance.
    
    Returns:
    - float: The variance of the input data.
    
    Requirements:
    - numpy
    
    Example:
    >>> var = get_var([1, 2, 3, 4, 5])
    >>> print(var)
    2.0
    """
    # Calculate the mean of the data
    mean = np.mean(data)
    
    # Calculate the variance using the formula: sum((x - mean)^2) / n
    var = sum([np.power(x - mean, 2) for x in data]) / len(data)
    
    return var
\end{spverbatim}
        \end{minipage}\\
    \bottomrule
    \end{tabular}
    \caption{Demo of the code snippet after refinement.}
    \label{tab:appendix-generator-prompt}
\end{table*}

\begin{table*}[htbp]
    \centering
    \small
    \begin{tabular}{l}
    \toprule
    \textbf{Prompt for the unit test generator} \\
    \midrule
    \begin{minipage}{0.8\textwidth}
\begin{spverbatim}
You are a professional software testing expert. Your task is to write comprehensive unit tests for the given function.

Please follow these guidelines:
1. Write tests that cover different scenarios including:
   - Normal/expected inputs
   - Edge cases
   - Invalid/unexpected inputs
   - Boundary conditions

2. Each test case should:
   - Have a clear and descriptive name
   - Include assertions that verify both return values and expected behavior
   - Be independent of other test cases
   - Include brief comments explaining the test purpose

3. Test structure requirements:
   - Use the unittest framework
   - Create a proper test class inheriting from unittest.TestCase
   - Include setUp/tearDown methods if necessary
   - Write self-contained tests that don't rely on external resources

4. Important:
   - Only output the test code within Python code blocks
   - Ensure all necessary imports are included
   - Focus on functionality testing rather than implementation details
   - Write tests that are maintainable and readable

Please analyze the given function and generate appropriate unit tests following these guidelines.
Your output format should be like this:
```python
import unittest
class TestCases(unittest.TestCase):
    def test_case_1(self):
        # Test purpose: Verify the function handles normal inputs correctly
        self.assertEqual(function_name(input1, input2), expected_output1)
    def test_case_2(self):
        ... 
```
Do not modify the class name(TestCases).
\end{spverbatim}
    \end{minipage}\\
    \bottomrule
    \end{tabular}
    \caption{Prompt for unit test generator.}
    \label{tab:appendix-prompt-template-unit-test-generator}
\end{table*}

\begin{table*}[htbp]
    \centering
    \small
    \begin{tabular}{l}
    \toprule
    \textbf{Prompt for the bug-fix agent.} \\
    \midrule
    \begin{minipage}{0.8\textwidth}
\begin{spverbatim}
You are a powerful coding expert specialized in code debugging and optimization. Your task is to fix the given code based on unit test results and error messages.

Please follow these guidelines:
1. Carefully analyze:
   - The original code implementation
   - Failed test cases and their error messages
   - Test requirements and expected behavior

2. When fixing the code:
   - Make minimal necessary changes to fix the issues
   - Maintain the original code structure when possible
   - Ensure the solution is efficient and clean

3. Important:
   - Only output the fixed code within Python code blocks
   - Ensure the solution passes all test cases
   - Focus on addressing the specific test failures
   - Maintain code readability and best practices

Please analyze the code and test failures, then provide the corrected implementation.
Your output format should be like this:
```python
# imports
def function_name(params):
    # Fixed implementation
    ...
```
\end{spverbatim}
    \end{minipage}\\
    \bottomrule
    \end{tabular}
    \caption{Prompt for bug-fix agent.}
    \label{tab:appendix-prompt-template-unit-test-generator}
\end{table*}

\begin{table*}[htbp]
    \centering
    \small
    \begin{tabular}{l}
    \toprule
    \textbf{Prompt for the refine agent.} \\
    \midrule
    \begin{minipage}{0.8\textwidth}
\begin{spverbatim}
You are a powerful coding expert specialized in code documentation and optimization. 
Given a code snippet and its unit tests, please enhance the code with comprehensive documentation while maintaining its functionality.

Requirements:
1. Documentation Enhancement:
   - Add clear function description
   - Document parameters and return values
   - List required dependencies
   - Provide usage examples
   - Document potential exceptions (if applicable)

2. Code Refinement Guidelines:
   - Add concise inline comments at key points
   - Maintain code functionality
   - Ensure code remains readable and well-styled
   - Add necessary error handling without affecting core logic
   - Keep function names unchanged

3. Documentation Format:
   - Function description
   - Parameters
   - Returns
   - Requirements
   - Raises (if applicable)
   - Examples

Your output should follow this structure:
```python
def function_name(params):
    # Core documentation
    # Implementation with inline comments
    ...
```
\end{spverbatim}
    \end{minipage}\\
    \bottomrule
    \end{tabular}
    \caption{Prompt for refine agent.}
    \label{tab:appendix-prompt-template-refine}
\end{table*}

\end{document}